\documentclass{article}
\usepackage[T1]{fontenc}

\usepackage{microtype}
\usepackage{graphicx}
\usepackage{subcaption}
\usepackage{booktabs}
\usepackage{multirow}
\usepackage{xcolor}
\usepackage{hyperref}

\usepackage[preprint]{icml2026}

\usepackage{amsmath}
\usepackage{amssymb}
\usepackage{mathtools}
\usepackage{amsthm}
\usepackage[capitalize,noabbrev]{cleveref}

\newcommand{\nVniah}{720}
\newcommand{\nAbl}{240}
\newcommand{\accUni}{52.2}

\newcommand{\accUniSixty}{38.8}
\newcommand{\accMMEUni}{55.6}
\newcommand{\accEditFull}{49.6}

\newcommand{\accMMEEditFull}{53.9}

\newcommand{\accEditScr}{50.7}

\newcommand{\accMMEEditScr}{54.4}
\newcommand{\accOracleFull}{96.9}

\newcommand{\accOracleFullSixty}{91.7}

\newcommand{\seenInit}{37.1}
\newcommand{\seenInitTwenty}{62.9}
\newcommand{\seenInitSixty}{17.5}
\newcommand{\locRecall}{29.2}

\newcommand{\locRecallUnseen}{6.8}
\newcommand{\locRecallSeen}{67.0}
\newcommand{\accEditFullLocOK}{97.6}
\newcommand{\nLocOK}{210}
\newcommand{\gpuHoursVniah}{9.3}
\newcommand{\gpuHoursBudget}{18.5}
\newcommand{\selModeSeg}{17}
\newcommand{\selModeBudget}{76.7}
\newcommand{\seenInitBudget}{67.8}
\newcommand{\locRecallBudget}{7.5}
\newcommand{\nBudget}{720}
\newcommand{\accUniBudget}{63.1}
\newcommand{\accEditFullBudget}{53.9}
\newcommand{\accOracleFullBudget}{100.0}
\newcommand{\accUniMatched}{52.2}

\newcommand{\agreeScrBudget}{94.3}
\newcommand{\tokSaveFullBudget}{34}
\newcommand{\tokInitBudget}{46{,}077}
\newcommand{\gpuHoursMME}{1.3}
\newcommand{\accUniSeen}{100.0}
\newcommand{\accUniUnseen}{24.1}
\newcommand{\parsePlain}{81.4}
\newcommand{\parseForced}{99.6}

\newcommand{\tokFull}{17{,}361}
\newcommand{\latFull}{7.20}
\newcommand{\llmLatFull}{1.46}

\newcommand{\flopFull}{416}

\newcommand{\tokPrefix}{16{,}571}

\newcommand{\llmLatPrefix}{2.56}

\newcommand{\tokScr}{12{,}191}
\newcommand{\latScr}{4.26}
\newcommand{\llmLatScr}{1.67}

\newcommand{\flopScr}{263}

\newcommand{\tokInit}{23{,}549}
\newcommand{\tokEdit}{17{,}361}

\newcommand{\relocTok}{4{,}380}

\newcommand{\deltaFail}{0}
\newcommand{\tokSaveFull}{30}

\newcommand{\flopSaveFull}{37}
\newcommand{\latSaveFull}{41}

\newcommand{\speedFull}{1.7}

\newcommand{\vitSave}{63}
\newcommand{\klScr}{0.0049}

\newcommand{\agreeScr}{97.4}

\newcommand{\agreeOracleScr}{99.9}

\newcommand{\agreePrefix}{98.5}
\newcommand{\klScrMME}{0.0062}
\newcommand{\agreeScrMME}{96.1}
\newcommand{\tokSaveFullMME}{32}
\newcommand{\speedFullMME}{1.8}
\newcommand{\nMME}{180}
\newcommand{\nMMEvid}{60}
\newcommand{\durMME}{41}
\newcommand{\accLVBUni}{56.2}
\newcommand{\accLVBEditFull}{55.4}

\newcommand{\accLVBEditScr}{55.1}
\newcommand{\agreeScrLVB}{97.1}
\newcommand{\tokSaveLVB}{32}
\newcommand{\latFullLVB}{6.23}
\newcommand{\latScrLVB}{3.49}
\newcommand{\speedLVB}{1.8}
\newcommand{\gpuHoursLVB}{1.9}

\newcommand{\agreeScrLVBQ}{98.9}
\newcommand{\tokSaveLVBQ}{29}
\newcommand{\latFullLVBQ}{2.02}

\newcommand{\speedLVBQ}{1.2}
\newcommand{\gpuHoursLVBQ}{0.8}

\newcommand{\agreeScrLVBT}{94.6}
\newcommand{\tokSaveLVBT}{33}

\newcommand{\speedLVBT}{1.2}
\newcommand{\gpuHoursLVBT}{0.8}
\newcommand{\klScrLVB}{0.0051}
\newcommand{\nLVB}{276}
\newcommand{\nLVBvid}{92}
\newcommand{\durLVB}{31}
\newcommand{\fiveOptLVB}{86}
\newcommand{\nInfra}{90}
\newcommand{\infraTokSix}{30}

\newcommand{\infraKLSix}{0.0081}

\newcommand{\infraTokAll}{63}

\newcommand{\infraSpeedAll}{1.98}
\newcommand{\infraSpeedFAAll}{2.29}
\newcommand{\infraKLAll}{0.0242}
\newcommand{\infraAgreeAll}{94.9}
\newcommand{\infraKZeroAll}{0.0032}

\newcommand{\infraLatFAFull}{3.49}
\newcommand{\infraLatFAAll}{1.52}
\newcommand{\nInfraMME}{28}
\newcommand{\infraTokAllMME}{64}
\newcommand{\infraSpeedAllMME}{2.09}
\newcommand{\infraKLAllMME}{0.0267}
\newcommand{\infraKLSixMME}{0.0111}
\newcommand{\nInfraLVB}{24}
\newcommand{\infraTokAllLVB}{62}
\newcommand{\infraSpeedAllLVB}{2.00}
\newcommand{\infraKLAllLVB}{0.0338}
\newcommand{\infraKLSixLVB}{0.0112}
\newcommand{\ablKOnerZeroKL}{0.0009}
\newcommand{\ablKOnerZeroAgree}{98.8}

\newcommand{\ablKSixrZeroKL}{0.0082}

\newcommand{\ablKTwelverZeroKL}{0.0142}
\newcommand{\ablKTwelverZeroAgree}{96.7}
\newcommand{\ablKSixrSixteenKL}{0.0079}

\newcommand{\ablKSixrSixtyFourKL}{0.0082}

\newcommand{\testRelErr}{0.0030}
\newcommand{\testAbsErr}{1.00}
\newcommand{\testMaxKey}{174}
\newcommand{\testNoShift}{0.055}
\newcommand{\testNoEditKL}{9.4e-07}
\newcommand{\testPrefixKL}{0.0000}
\newcommand{\testSegs}{19}
\newcommand{\nFA}{720}
\newcommand{\latFAFull}{3.43}
\newcommand{\llmLatFAFull}{1.44}
\newcommand{\latSDFull}{7.20}
\newcommand{\llmLatSDFull}{1.46}
\newcommand{\latFAPrefix}{3.29}
\newcommand{\llmLatFAPrefix}{1.39}
\newcommand{\latSDPrefix}{8.09}
\newcommand{\llmLatSDPrefix}{2.56}
\newcommand{\latFAScr}{1.88}
\newcommand{\llmLatFAScr}{1.13}
\newcommand{\latSDScr}{4.26}
\newcommand{\llmLatSDScr}{1.67}
\newcommand{\speedFA}{1.8}
\newcommand{\llmSaveFA}{21}
\newcommand{\agreeScrFA}{97.4}
\newcommand{\nAppend}{720}
\newcommand{\accAppend}{55.7}

\newcommand{\accOracleAppend}{97.4}
\newcommand{\accEditScrAppendSet}{50.7}
\newcommand{\accOracleFullAppendSet}{96.9}

\newcommand{\tokAppend}{6{,}303}
\newcommand{\latAppend}{3.98}
\newcommand{\ctxAppend}{29{,}794}
\newcommand{\tokScrAppendSet}{12{,}191}
\newcommand{\latScrAppendSet}{4.26}
\newcommand{\ctxEditAppendSet}{17{,}361}

\newcommand{\nQ}{720}
\newcommand{\accQUni}{55.6}

\newcommand{\accQEditFull}{53.3}

\newcommand{\accQEditScr}{54.4}

\newcommand{\accQOracleFull}{96.8}

\newcommand{\agreeQScr}{95.8}

\newcommand{\tokSaveQ}{34}

\newcommand{\locRecallQ}{32.2}
\newcommand{\locRecallUnseenQ}{2.6}

\newcommand{\accTUni}{52.8}

\newcommand{\accTEditFull}{52.4}

\newcommand{\accTEditScr}{52.4}

\newcommand{\accTOracleFull}{96.8}

\newcommand{\agreeTScr}{97.2}

\newcommand{\tokSaveT}{31}

\newcommand{\locRecallT}{38.2}
\newcommand{\locRecallUnseenT}{4.4}

\newcommand{\testRelErrQ}{0.0007}
\newcommand{\testNoShiftQ}{0.068}
\newcommand{\testRelErrT}{0.0009}

\newcommand{\gpuHoursFA}{4.5}
\newcommand{\gpuHoursAppend}{7.5}
\newcommand{\gpuHoursQ}{3.5}
\newcommand{\gpuHoursQMME}{0.6}
\newcommand{\gpuHoursT}{3.9}
\newcommand{\gpuHoursTMME}{0.6}

\newcommand{\scr}{SCR-VL}

\icmltitlerunning{Editable Video Context with Mid-Context KV Reuse}

\begin{document}

\twocolumn[
  \icmltitle{Editable Video Context: Mid-Context KV-Cache Reuse\\for Context-Managing Video LLMs}

  \icmlsetsymbol{equal}{*}

  \begin{icmlauthorlist}
    \icmlauthor{Dayu Wang}{equal,ntu,baidu}
    \icmlauthor{Shasha Xu}{equal,pku}
  \end{icmlauthorlist}

  \icmlaffiliation{ntu}{Nanyang Technological University, Singapore}
  \icmlaffiliation{baidu}{Baidu Inc., Beijing, China}
  \icmlaffiliation{pku}{Peking University, Beijing, China}

  \icmlcorrespondingauthor{Dayu Wang}{dayu001@e.ntu.edu.sg}
  \icmlcorrespondingauthor{Shasha Xu}{elsa031218@stu.pku.edu.cn}

  \icmlkeywords{Video LLM, KV cache, context management, M-RoPE}

  \vskip 0.3in
]

\printAffiliationsAndNotice{\icmlEqualContribution}

\begin{abstract}
Context Language Models let a language model edit its own context like a file, and Suffix Cache Reuse makes such
mid-context edits cheap for text by re-rotating the cached keys of surviving tokens instead of recomputing them. We
carry both ideas to video VLMs with multimodal RoPE (M-RoPE). A long video is held as a list of timestamped segments;
the model first watches a sparse version, names the segments it needs, and the context is edited so those segments are
re-decoded at 4$\times$ frame rate while others are deleted. Our rebuild method, \scr{}, reuses the KV of surviving
segments by rotating their keys by one offset $\Delta$, which we show (and assert per segment) is shared by the
temporal, height and width components of Qwen2.5-VL's M-RoPE; a content-hashed cache of vision-encoder outputs removes
ViT work for survivors. In \emph{preliminary} experiments with Qwen2.5-VL-7B on \nVniah{} synthetic
needle-in-a-video-haystack questions over public-domain films, \scr{} cuts prefill tokens by \tokSaveFull\% and
end-to-end rebuild latency by \latSaveFull\% relative to full recomputation while agreeing with its answer on
\agreeScr\% of questions. On \nLVB{} LongVideoBench questions over videos of at least 20 minutes it agrees with full
recomputation on \agreeScrLVB\% of questions at \speedLVB$\times$ lower rebuild latency, and relocating every survivor
instead of the 6 largest raises the saving to \infraTokAll\% of prefill tokens (\infraSpeedFAAll$\times$ with
FlashAttention-2) in a smaller answer-free fidelity study. Editing helps when the edit is right: densifying the true segment raises accuracy from
\accUni\% to \accOracleFull\%. When the model chooses the segments itself, accuracy does not improve
(\accEditFull\%), because it rarely finds a needle that the sparse pass did not already sample. Appending the
densified segments instead of editing in place is a stronger and cheaper baseline in this setting
(\accAppend\% accuracy, \tokAppend{} prefilled tokens). The same pattern holds for Qwen3-VL-8B Instruct and Thinking.
Code and aggregated results are available at \url{https://github.com/dayuwang401/SCR-VL}.
\end{abstract}

\section{Introduction}
\label{sec:intro}

Long videos do not fit a VLM's context at useful frame rates. Uniform sparse sampling is the default, and it fails in
a predictable way: a few-second event in an hour of footage is often missed, or seen in a single blurred frame
\citep{fu2024videomme,zhou2024mlvu,wu2024longvideobench}. Agentic systems instead retrieve frames on demand through
external tools \citep{wang2024videoagent,wang2025videotree,zhang2025deepvideodiscovery}, and streaming systems keep or
retrieve KV caches as the video arrives \citep{di2025rekv,xu2026streamingvlm}.

Context Language Models (CLMs; \citealp{shao2026clm}) suggest a simpler interface: let the model manage its own
context by editing it freely. Edits in the middle of a context are expensive for standard serving, which only reuses
the KV cache up to the first changed token \citep{zheng2023sglang,kwon2023pagedattention}. CLM therefore introduces
Suffix Cache Reuse (SCR): tokens that survive after an edit keep their KV, and their keys are rotated to the new
positions, which is exact for RoPE \citep{su2021roformer} at the first layer and approximate deeper in the network.
SCR targets text models with 1D RoPE.

\begin{figure*}[t]
  \centering
  \includegraphics[width=\textwidth]{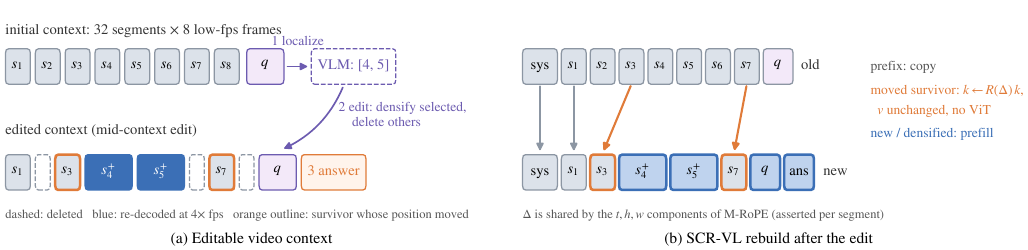}
  \caption{\textbf{Editable video context and \scr{}.} (a) The VLM sees 32 sparse segments and the question, names
  the segments it needs (step 1), and the context is edited mid-sequence: selected segments are re-decoded at
  4$\times$ frame rate, other even-numbered segments are deleted (step 2), and the model answers on the edited context
  (step 3). (b) Rebuilding the KV cache after the edit: the unchanged prefix is copied; surviving segments whose
  position moved are reused by rotating their keys by $\Delta$ (values unchanged, no ViT pass); new and densified
  segments are prefilled on top of the rebuilt cache.}
  \label{fig:overview}
\end{figure*}

This paper asks whether the same recipe works for video VLMs, where (i) the context is dominated by visual tokens,
(ii) positions are three-dimensional (M-RoPE; \citealp{wang2024qwen2vl,bai2025qwen25vl}), and (iii) a vision encoder
adds a second, large cost to every recomputation. We hold a long video as a list of timestamped segments and let the
model edit that list (\cref{fig:overview}a): after a sparse first look, it names the segments it needs, which are then
re-decoded at a higher frame rate while other segments are dropped. Survivors after the first edit change position;
\scr{} moves their KV instead of recomputing it (\cref{fig:overview}b).

\textbf{Contributions.}
\begin{enumerate}
\item \textbf{Editable video context.} A two-round localize-then-densify protocol for long-video QA in which the edit
lands mid-context, so prefix caching alone cannot avoid recomputing the survivors after it.
\item \textbf{\scr{}.} When a segment moves as a whole, all three M-RoPE components shift by the same $\Delta$, so a
single rotation $R(\Delta)$ of the cached key is exact at layer 0 (relative error \testRelErr{} in bf16, versus
\testNoShift{} without the rotation). A content-hashed ViT cache makes survivors free on the vision side.
\item \textbf{A measured, partly negative evaluation.} On \nVniah{} V-NIAH questions, \scr{} reduces rebuild
compute by \flopSaveFull\% and latency by \latSaveFull\% at \agreeScr\% answer agreement with full recompute. Oracle
edits show the edited context itself is valuable (\accUni\%$\rightarrow$\accOracleFull\%), but Qwen2.5-VL-7B's own
localization is not good enough to realize that gain. We also report that the LLM-side speedup does not show up in our
HuggingFace implementation (\cref{sec:limits}).
\end{enumerate}

\section{Related Work}
\label{sec:related}

\textbf{KV reuse beyond the prefix.} Prefix caching shares KV across requests with a common prefix
\citep{zheng2023sglang,kwon2023pagedattention}. Prompt Cache reuses attention states of precomputed prompt modules
at fixed position ranges \citep{gim2024promptcache}; CacheBlend fuses cached KV of retrieved chunks and recomputes a
small set of tokens to restore cross-chunk attention \citep{yao2024cacheblend}; EPIC caches chunks independently of
position and recomputes a few boundary tokens \citep{hu2024epic}. SCR in CLMs moves the surviving suffix after a
model-made edit \citep{shao2026clm}. Our boundary-repair variant ($r>0$) follows the recompute-a-few-tokens idea of
CacheBlend and EPIC. All of these target 1D RoPE text models; we handle M-RoPE and the vision encoder.

\textbf{Long and streaming video.} Long-context VLMs extend the visual token budget
\citep{zhang2024longva,shu2024videoxl,qin2025videoxl2}; LongVA also introduced visual needle-in-a-haystack tests,
which our synthetic task follows in spirit. StreamingLLM keeps attention sinks plus a recent window
\citep{xiao2024streamingllm}; ReKV retrieves stored video KV for streaming QA \citep{di2025rekv}; StreamingVLM targets
real-time understanding of unbounded streams \citep{xu2026streamingvlm}. These methods decide what to keep with fixed
rules or retrieval; here the model edits its own context.

\textbf{Agents over video.} VideoAgent \citep{wang2024videoagent}, VideoTree \citep{wang2025videotree} and Deep Video
Discovery \citep{zhang2025deepvideodiscovery} gather frames iteratively with an LLM controller. Our two-round
protocol is a minimal in-context version of this loop; the contribution is making each edit cheap to re-serve.

\section{Method}
\label{sec:method}

\subsection{Editable video context}
A video of duration $D$ is split into $N{=}32$ equal segments. Segment $i$ is a line of text
``\texttt{Segment $i$ <t=hh:mm:ss-hh:mm:ss>:}'' followed by its own video block of $n_i$ frames sampled uniformly from
its time span, each block encoded separately by the ViT. The context is
$[\text{system}]\,[s_1]\cdots[s_N]\,[q]$, where $q$ is the question and options, placed last so that the segments'
KV does not depend on it. Initially $n_i{=}8$ at $448{\times}308$ pixels (176 visual tokens per pair of frames), giving
\tokInit{} tokens, about 72\% of a 32K budget.

\textbf{Round 1 (localize).} We append ``Which segments most likely contain the frames needed to answer the question?
Reply only with a JSON list of at most 3 segment numbers'' and decode greedily. With an unconstrained reply, Qwen2.5-VL-7B
produced a parsable non-empty list on \parsePlain\% of V-NIAH questions (almost all failures were the empty list
\texttt{[]}), so we pre-fill the reply with ``\texttt{[}''. This parsed on \parseForced\% of questions; the rest were
lists cut off by the 24-token limit, and those questions keep the unedited context.

\textbf{Edit.} Each selected segment is replaced by the same time span re-decoded with $n_i{=}32$ frames; unselected
segments with an even index are deleted. This rule is fixed and mid-context: segment 2 is deleted unless selected, so
every survivor after $s_1$ changes position. The edited context averages \tokEdit{} tokens.

\textbf{Round 2 (answer).} The model answers on the edited context; we read the logits of the option letters at the
first answer position.

\subsection{Rebuilding the cache: full, prefix and \scr{}}
Given the old cache (built for the initial context) and the edited segment list, \textbf{full} prefills the whole new
sequence including all ViT passes; \textbf{prefix} copies the KV up to the first block whose content hash differs and
prefills the rest. Both are exact up to floating-point order. \textbf{\scr{}} copies the same prefix, then walks the
remaining blocks: a surviving segment (same content hash as in the old context) is \emph{relocated} by copying its KV
from the old cache and rotating its keys; any other block is prefilled on top of the cache built so far, using its
correct M-RoPE positions in the new sequence. Following CLM, at most $K$ survivors are relocated (largest first,
default $K{=}6$) and the rest are prefilled. In the $r>0$ variant, the first $r$ tokens of each relocated segment are
recomputed against the new left context and only the remainder is copied.

\subsection{M-RoPE shift}
Qwen2.5-VL assigns each token a position triple $(p^t,p^h,p^w)$. Text tokens use $p^t{=}p^h{=}p^w$; a video block
starting after position $s$ gets $p^t = s + \lfloor \tau\, g\, j\rfloor$, $p^h = s + y$, $p^w = s + x$ for temporal patch
$j$ and merged-patch coordinates $(y,x)$, where $\tau$ is seconds per temporal patch and $g$ the model's tokens per
second; the next block starts at one past the maximum of all three components \citep{wang2024qwen2vl,bai2025qwen25vl}.
Because each segment is its own block, its positions are a fixed local pattern plus the position of its first token.
Moving a segment therefore adds the same integer $\Delta$ to all three components of all its tokens.

The rotary embedding splits the 64 frequency pairs of a 128-dimensional head into sections of 16/24/24 pairs driven by
$p^t,p^h,p^w$. The cache stores rotated keys $\tilde k = R(\mathbf p)\,k$, and rotations compose, so
\begin{equation}
  R(\mathbf p + \Delta\mathbf 1)\,k = R(\Delta\mathbf 1)\,\tilde k ,
  \label{eq:shift}
\end{equation}
where $R(\Delta\mathbf 1)$ rotates frequency pair $i$ by $\Delta\theta_i$ irrespective of its section. We build the
cos/sin tables with the model's own inverse frequencies and section layout, apply \cref{eq:shift} in fp32, and leave
values unchanged. Before relocating a segment we compute the new positions with the model's \texttt{get\_rope\_index} on
the whole new sequence and assert that new minus old positions is one constant over all components and tokens; a
segment that fails the check would be recomputed (this never happened: \deltaFail{} failures). \Cref{eq:shift} is
exact at layer 0, where keys depend only on the token's own embedding. In deeper layers the copied KV still reflects
attention to the old left context, so \scr{} is an approximation, as in CLM.

\textbf{ViT cache.} Each segment's ViT output is cached under a SHA-256 of (video file and size, absolute start/end
time, frame count, resolution, needle specification if a sampled frame shows it, processor configuration). The cache is
exact. In our runs, full and prefix re-run the ViT for every segment they prefill, while \scr{} reuses cached outputs
for all survivors, including those beyond $K$ that it prefills.

\section{Experiments}
\label{sec:exp}

\subsection{Setup}
\textbf{Model and implementation.} Qwen2.5-VL-7B-Instruct \citep{bai2025qwen25vl} in bf16 with HuggingFace
transformers 4.51.3, PyTorch 2.7 and SDPA attention, one model per A800-80GB GPU. All three rebuild paths call the same
language-model forward on the same segment embeddings; they differ only in which blocks are prefilled and which KV is
copied. Latency is wall-clock after \texttt{torch.cuda.synchronize}, includes ViT and frame loading, and is measured
after one untimed warm-up per worker. Theoretical FLOPs are $2P_{\text{LLM}}T + 4LdA + 2P_{\text{ViT}}V$, with $T$
prefilled tokens, $A=\sum_{\text{prefilled }i} c_i$ the number of attended key positions ($L{=}28$ layers, $d{=}3584$),
and $V$ ViT patches.

\textbf{V-NIAH.} Haystacks are three public-domain films from the Internet Archive (\emph{Anna Christie}, 1930;
\emph{Disraeli}, 1929; \emph{Bulldog Drummond}, 1929; all Public Domain Mark). Each question takes a random 20-, 40- or
60-minute window and overwrites 2--4\,s at a uniformly random point in 10--90\% of the window with a full-screen card
showing one word (a solid colour with large text). The question asks which of four words was shown. We use 80
questions per film and length (\nVniah{} total); the \emph{oracle} edit densifies the segment(s) that overlap the
card.

\textbf{Video-MME.} The first 60 \emph{long} videos (by YouTube id) of Video-MME \citep{fu2024videomme}, fetched from the
dataset's official HuggingFace release for research use, \nMME{} questions, mean length \durMME{} minutes. No ground-truth
interval is available, so there is no oracle row.

\textbf{LongVideoBench.} All \nLVB{} validation questions of LongVideoBench \citep{wu2024longvideobench} whose video
is at least 20 minutes long (\nLVBvid{} videos, mean length \durLVB{} minutes; \fiveOptLVB\% of the questions have five
options, so chance is 20--25\%). We use the same pipeline and frame bank settings as for Video-MME, without subtitles
and without an oracle row.

\textbf{Conditions.} \emph{uniform}: answer on the initial context. \emph{edit-full / edit-prefix / edit-\scr}: model
edit with the three rebuild paths. \emph{oracle-full / oracle-\scr}: the oracle edit. Confidence intervals are 95\%
bootstrap intervals over questions (2{,}000 resamples).

\subsection{Is \scr{} faithful?}
All five correctness tests pass (\cref{app:tests}). Relocated layer-0 keys match a full prefill of the new sequence
with maximum relative error \testRelErr{} (maximum absolute error \testAbsErr{} at key magnitudes up to
\testMaxKey{}, i.e., one bf16 step), while copying keys without the rotation gives at least \testNoShift{}.
End-to-end, \scr{} picks the same option as full recomputation on \agreeScr\% of V-NIAH questions (\agreeOracleScr\%
under the oracle edit), with mean KL(full$\,\|\,$\scr) of \klScr{} nats on the full-vocabulary next-token
distribution (\cref{tab:agree}). Prefix reuse, which is exact up to floating-point order, agrees on \agreePrefix\%,
so part of the disagreement is bf16 noise from chunked prefill rather than the approximation. Accuracy of edit-\scr{}
(\accEditScr\%) is within the confidence interval of edit-full (\accEditFull\%; \cref{tab:acc}).

\begin{table*}[t]
  \centering
  \caption{\textbf{Accuracy (\%)} with 95\% bootstrap CIs. Chance is 25\% (20--25\% on LongVideoBench). Edit rows use the model's own segment
  choice; oracle rows densify the segment containing the needle.}
  \label{tab:acc}
  \small
  \begin{tabular}{lcccccc}
\toprule
 & \multicolumn{4}{c}{V-NIAH (haystack length)} & Video-MME & LongVideoBench \\
\cmidrule(lr){2-5}\cmidrule(lr){6-6}\cmidrule(lr){7-7}
Condition & 20 min & 40 min & 60 min & all & long & $\geq$20 min \\
 & ($n$=240) & ($n$=240) & ($n$=240) & ($n$=720) & ($n$=180) & ($n$=276) \\
\midrule
uniform (no edit) & 70.8\,{\scriptsize[65,77]} & 47.1\,{\scriptsize[41,53]} & 38.8\,{\scriptsize[33,45]} & 52.2\,{\scriptsize[49,56]} & 55.6\,{\scriptsize[48,63]} & 56.2\,{\scriptsize[51,62]} \\
edit + full recompute & 70.4\,{\scriptsize[65,76]} & 42.1\,{\scriptsize[36,48]} & 36.2\,{\scriptsize[30,42]} & 49.6\,{\scriptsize[46,53]} & 53.9\,{\scriptsize[46,61]} & 55.4\,{\scriptsize[49,62]} \\
edit + prefix reuse & 70.4\,{\scriptsize[65,76]} & 42.5\,{\scriptsize[37,49]} & 35.8\,{\scriptsize[30,42]} & 49.6\,{\scriptsize[46,53]} & 54.4\,{\scriptsize[47,62]} & 56.2\,{\scriptsize[50,62]} \\
edit + SCR-VL (ours) & 71.2\,{\scriptsize[66,77]} & 44.2\,{\scriptsize[38,50]} & 36.7\,{\scriptsize[30,42]} & 50.7\,{\scriptsize[47,54]} & 54.4\,{\scriptsize[47,62]} & 55.1\,{\scriptsize[49,61]} \\
\midrule
oracle edit + full & 100.0\,{\scriptsize[100,100]} & 99.2\,{\scriptsize[98,100]} & 91.7\,{\scriptsize[88,95]} & 96.9\,{\scriptsize[96,98]} & -- & -- \\
oracle edit + SCR-VL & 100.0\,{\scriptsize[100,100]} & 99.2\,{\scriptsize[98,100]} & 91.7\,{\scriptsize[88,95]} & 96.9\,{\scriptsize[96,98]} & -- & -- \\
\bottomrule
\end{tabular}

\end{table*}

\begin{table}[t]
  \centering
  \caption{\textbf{Cost per question on V-NIAH} (means over \nVniah{} questions). ``initial'' builds the first
  context; other rows rebuild the cache after the edit. Moved tokens are copied and re-rotated by \scr{} instead of
  prefilled. LLM-only latency excludes ViT time.}
  \label{tab:cost}
  \setlength{\tabcolsep}{3pt}
  \resizebox{\columnwidth}{!}{\begin{tabular}{lrrrrrrr}
\toprule
Rebuild path & prefill & moved & ViT & TFLOPs & latency & LLM-only & mem \\
 & tokens & tokens & frames & & (s) & (s) & (GB) \\
\midrule
initial context & 23{,}549 & -- & 256 & 592 & 8.86 & 2.14 & 20.8 \\
\midrule
full recompute & 17{,}361 & 0 & 190 & 416 & 7.20 & 1.46 & 20.1 \\
prefix reuse & 16{,}571 & 0 & 182 & 399 & 8.09 & 2.56 & 21.5 \\
SCR-VL & 12{,}191 & 4{,}380 & 70 & 263 & 4.26 & 1.67 & 19.7 \\
\midrule
oracle full & 14{,}455 & 0 & 158 & 337 & 5.67 & 1.18 & 20.5 \\
oracle SCR-VL & 9{,}281 & 4{,}398 & 34 & 186 & 2.42 & 1.16 & 19.2 \\
\bottomrule
\end{tabular}
}
\end{table}

\subsection{What does \scr{} save?}
\begin{figure}[t]
  \centering
  \includegraphics[width=\columnwidth]{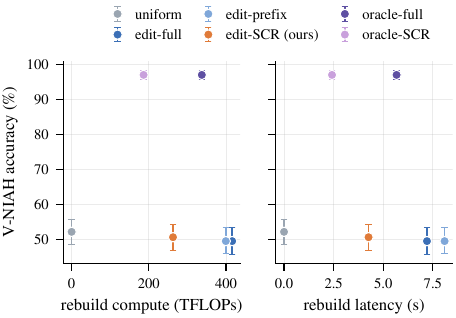}
  \caption{\textbf{Accuracy vs.\ rebuild cost} on V-NIAH (95\% CIs). \scr{} matches the accuracy of its exact
  counterpart at lower compute and latency, for both model and oracle edits. Uniform has no rebuild.}
  \label{fig:pareto}
\end{figure}

\Cref{tab:cost} and \cref{fig:pareto} give the costs. Relative to full recomputation, \scr{} prefills
\tokSaveFull\% fewer tokens (\tokScr{} vs.\ \tokFull{}; \relocTok{} tokens are moved instead), cuts theoretical
compute by \flopSaveFull\% (\flopScr{} vs.\ \flopFull{} TFLOPs) and runs \vitSave\% fewer ViT frames. End-to-end
rebuild latency falls from \latFull{}\,s to \latScr{}\,s (\speedFull$\times$). Prefix reuse saves almost nothing
(\tokPrefix{} tokens) because our edit changes the context right after $s_1$; this is the case SCR is designed for.

The latency gain comes from the vision side. LLM-only time is \llmLatScr{}\,s for \scr{} versus \llmLatFull{}\,s for
full recomputation, despite fewer prefilled tokens, and prefix reuse is slower still (\llmLatPrefix{}\,s). In our
implementation a full prefill runs one causal SDPA call, whereas prefilling on top of an existing cache needs an
explicit mask and, for \scr{}, several chunks between relocated segments; we did not profile this further. To test this, we re-ran all \nFA{} questions with FlashAttention-2 for both the
ViT and the LLM (\cref{tab:fa2}). With FlashAttention-2, \scr{} is also faster on the LLM side (\llmLatFAScr{}\,s vs.\
\llmLatFAFull{}\,s for full recomputation, \llmSaveFA\% less), and end-to-end rebuild latency drops from \latFAFull{}\,s to
\latFAScr{}\,s (\speedFA$\times$). Answer agreement with full recomputation under FlashAttention-2 is \agreeScrFA\%.

\begin{table}[t]
  \centering
  \caption{\textbf{Agreement with full recomputation} on the same edited context: mean KL(full$\,\|\,$path) of the
  first answer token over the full vocabulary, top-1 token agreement, and agreement of the chosen option.}
  \label{tab:agree}
  \small
  \resizebox{\columnwidth}{!}{\begin{tabular}{lccc}
\toprule
Path vs.\ its full recompute & KL (nats) & top-1 token & answer \\
\midrule
prefix (V-NIAH) & 0.0006 & 98.5\% & 98.5\% \\
SCR-VL, model edit (V-NIAH) & 0.0049 & 97.4\% & 97.4\% \\
SCR-VL, oracle edit (V-NIAH) & 0.0001 & 99.9\% & 99.9\% \\
prefix (Video-MME) & 0.0007 & 97.8\% & 97.8\% \\
SCR-VL, model edit (Video-MME) & 0.0062 & 96.1\% & 96.1\% \\
\bottomrule
\end{tabular}
}
\end{table}

\begin{table}[t]
  \centering
  \caption{\textbf{Rebuild latency (s) with SDPA vs.\ FlashAttention-2} on the same \nFA{} V-NIAH questions.
  ``LLM'' excludes ViT time.}
  \label{tab:fa2}
  \small
  \begin{tabular}{lcccc}
    \toprule
     & \multicolumn{2}{c}{SDPA} & \multicolumn{2}{c}{FlashAttention-2} \\
    \cmidrule(lr){2-3}\cmidrule(lr){4-5}
    Rebuild path & total & LLM & total & LLM \\
    \midrule
    full recompute & \latSDFull & \llmLatSDFull & \latFAFull & \llmLatFAFull \\
    prefix reuse & \latSDPrefix & \llmLatSDPrefix & \latFAPrefix & \llmLatFAPrefix \\
    \scr{} & \latSDScr & \llmLatSDScr & \latFAScr & \llmLatFAScr \\
    \bottomrule
  \end{tabular}
\end{table}

\subsection{Does editing the context help?}
\begin{figure}[t]
  \centering
  \includegraphics[width=\columnwidth]{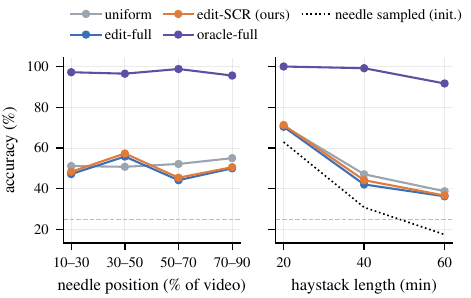}
  \caption{\textbf{V-NIAH accuracy by needle position (left) and haystack length (right).} The dotted line is the
  fraction of questions whose card is hit by one of the 256 initially sampled frames. The dashed grey line is chance.}
  \label{fig:position}
\end{figure}

\begin{figure*}[t]
  \centering
  \includegraphics[width=\textwidth]{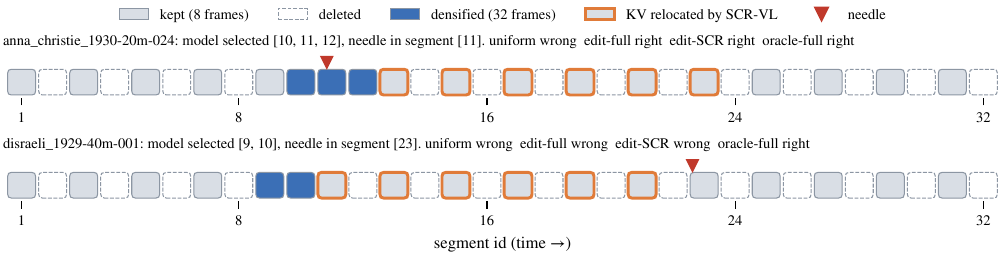}
  \caption{\textbf{Two edits.} Top: the model selects the needle segment and every condition except uniform answers
  correctly. Bottom: the selection misses the needle segment; only the oracle edit answers correctly from the
  densified context. Orange outlines mark the survivors \scr{} relocates ($K{=}6$); other kept segments after the first
  edit are prefilled.}
  \label{fig:timeline}
\end{figure*}

\textbf{Only when the edit is right.} Densifying the true segment lifts V-NIAH accuracy from \accUni\% to
\accOracleFull\% (\cref{tab:acc}), and the gain grows with haystack length (\cref{fig:position}, right): at 60 minutes,
uniform reaches \accUniSixty\% and oracle-full \accOracleFullSixty\%. The model's own edit does not help
(\accEditFull\% vs.\ \accUni\%). Uniform is near-perfect when one of its 256 frames lands on the card
(\accUniSeen\%) and near chance otherwise (\accUniUnseen\%); the card is sampled in \seenInit\% of questions
(\seenInitTwenty\% at 20 minutes, \seenInitSixty\% at 60). The model's selection contains the needle segment in
\locRecall\% of questions, \locRecallSeen\% when the sparse pass sampled the card but only \locRecallUnseen\% when it
did not. In other words, localization mostly succeeds where the answer was already visible, so the edit adds little.
When the selection is right ($n{=}$\nLocOK), edit-full is correct on \accEditFullLocOK\% of questions.
\Cref{fig:timeline} shows one success and one failure.

\textbf{Needle position.} Accuracy varies little with the needle's position in the video for any condition
(\cref{fig:position}, left); there is no lost-in-the-middle pattern at this scale.

\textbf{edit-append control.} Instead of editing in place, one can keep the initial context and append the densified
segments after the last one, just before the question; prefix reuse then covers the whole original context exactly. On
\nAppend{} V-NIAH questions with the same model selection, appending reaches \accAppend\% (\accEditScrAppendSet\% for
in-place edit-\scr{} on the same questions) and the oracle selection appended reaches \accOracleAppend\%
(\accOracleFullAppendSet\% in place). The appended context is longer (\ctxAppend{} vs.\ \ctxEditAppendSet{} tokens) because nothing is deleted,
but its rebuild is cheaper than in-place editing with \scr{}: \tokAppend{} vs.\ \tokScrAppendSet{} prefilled tokens and
\latAppend{}\,s vs.\ \latScrAppendSet{}\,s (SDPA). In this setting, where the edit only adds detail, appending is the
stronger baseline; in-place editing pays off only when the deleted content must leave the context, e.g.\ to stay
within a budget, which our 32K runs never hit.

\subsection{Ablations: $K$ and $r$}
\begin{figure}[t]
  \centering
  \includegraphics[width=\columnwidth]{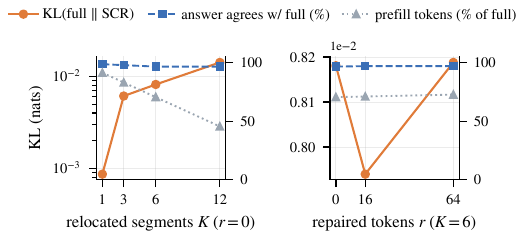}
  \caption{\textbf{Effect of $K$ (left) and boundary repair $r$ (right)} on KL to full recomputation (orange, left
  axis), option agreement and prefill tokens relative to full recomputation (right axis).}
  \label{fig:ablation}
\end{figure}

On the ablation subset (\nAbl{} questions; by construction of our sampling order these are all 20-minute haystacks,
so their accuracy is not comparable to \cref{tab:acc}), relocating more survivors lowers prefill tokens and latency and
raises the divergence from full recomputation (\cref{fig:ablation}, \cref{tab:ablation} in \cref{app:more}): KL grows
from \ablKOnerZeroKL{} at $K{=}1$ to \ablKTwelverZeroKL{} at $K{=}12$, while option agreement stays between
\ablKTwelverZeroAgree\% and \ablKOnerZeroAgree\%. Recomputing the first $r\in\{16,64\}$ tokens of each relocated
segment barely changed KL (\ablKSixrZeroKL{} at $r{=}0$, \ablKSixrSixteenKL{} at $r{=}16$, \ablKSixrSixtyFourKL{} at
$r{=}64$). The first 16 tokens of a segment are mostly its timestamp label; the error presumably sits in the visual
tokens, which this repair does not touch.

\textbf{Relocating every survivor.} Task accuracy is too noisy to rank rebuild variants (its 95\% CI on V-NIAH is about
$\pm$4 points), so we also built an answer-free harness (\cref{app:infra}). It freezes the edits, replays them through
each rebuild path, and compares the result with full recomputation on the probability of full recomputation's own
32-token greedy continuation (probe KL). On \nInfra{} V-NIAH questions, raising $K$ trades fidelity for compute
smoothly (\cref{tab:infra}). Relocating all survivors cuts prefill tokens by \infraTokAll\% (vs.\ \infraTokSix\% at
$K{=}6$) and reaches \infraSpeedAll$\times$ with SDPA and \infraSpeedFAAll$\times$ with FlashAttention-2
(\infraLatFAFull{}\,s$\rightarrow$\infraLatFAAll{}\,s). The cost is a probe KL of \infraKLAll{} vs.\ \infraKLSix{}, with
\infraAgreeAll\% of continuation tokens keeping the same argmax. Layer-0 keys of relocated tokens stay at one bf16
step from the exact ones (relative error \infraKZeroAll{}), so the extra error comes from deeper layers. $K$ is
therefore a knob rather than a fixed design choice: $K{=}6$ is the conservative default, and larger $K$ suits settings
that can tolerate more divergence from full recomputation.

\textbf{Larger budget.} Doubling the initial frames to 16 per segment (\tokInitBudget{} tokens, a 64K budget;
densified segments get 64 frames) on all \nBudget{} questions raises uniform accuracy to \accUniBudget\%
(\accUniMatched\% for the same questions at 32K), since the card is sampled in \seenInitBudget\% of them. Editing then
\emph{hurts}: edit-full reaches \accEditFullBudget\%, because the model's selection collapses to the same single
segment (segment \selModeSeg) on \selModeBudget\% of questions and contains the needle segment on only \locRecallBudget\%.
Oracle-full reaches \accOracleFullBudget\% (\cref{tab:budget}). \scr{} prefills \tokSaveFullBudget\% fewer tokens than
full recomputation at this budget and agrees with it on \agreeScrBudget\% of questions.

\subsection{Real long videos: Video-MME and LongVideoBench}
On \nMME{} long-video questions (\cref{tab:acc}), uniform reaches \accMMEUni\%, edit-full
\accMMEEditFull\% and edit-\scr{} \accMMEEditScr\%. \scr{} agrees with full recomputation on \agreeScrMME\% of
questions (mean KL \klScrMME{}) and prefills \tokSaveFullMME\% fewer tokens, with a \speedFullMME$\times$ lower rebuild
latency (\cref{tab:cost_mme} in \cref{app:more}). On the \nLVB{} LongVideoBench questions, uniform reaches
\accLVBUni\%, edit-full \accLVBEditFull\% and edit-\scr{} \accLVBEditScr\%. \scr{} agrees with full recomputation on
\agreeScrLVB\% of questions (mean KL \klScrLVB{}), prefills \tokSaveLVB\% fewer tokens and cuts rebuild latency from
\latFullLVB{}\,s to \latScrLVB{}\,s (\speedLVB$\times$; \cref{tab:cost_lvb}). Both samples are small and the CIs are
wide. We read them as a check that the conclusions from synthetic data hold on real videos: \scr{} tracks full
recomputation, and the model's own edit does not raise accuracy over uniform sampling.

\subsection{Other models: Qwen3-VL-8B Instruct and Thinking}
Qwen3-VL \citep{bai2025qwen3vl} differs in three ways that matter here: each temporal patch is its own vision block
preceded by a text timestamp, the M-RoPE frequencies are interleaved across $t,h,w$, and DeepStack adds ViT features to
the hidden states of the first decoder layers. A moved segment is still a pure shift (all five correctness tests pass;
layer-0 relative error \testRelErrQ{} vs.\ \testNoShiftQ{} without rotation), and with equal $\Delta$ on all
components the interleaving does not change the rotation. On \nQ{} V-NIAH questions (\cref{tab:models}), uniform
reaches \accQUni\%, edit-full \accQEditFull\%, edit-\scr{} \accQEditScr\% and oracle-full \accQOracleFull\%. \scr{}
agrees with full recomputation on \agreeQScr\% of questions and prefills \tokSaveQ\% fewer tokens. The selection
contains the needle segment on \locRecallQ\% of questions (\locRecallUnseenQ\% when the sparse pass missed it).
We also run Qwen3-VL-8B-Thinking with identical prompts; its chat template opens a reasoning block after the
assistant tag, which we close empty (\texttt{<think>\textbackslash n\textbackslash n</think>}) so that it answers directly
like the other models. Its tests pass as well (layer-0 relative error \testRelErrT{}). On V-NIAH it reaches \accTUni\%
uniform, \accTEditFull\% edit-full, \accTEditScr\% edit-\scr{} and \accTOracleFull\% oracle-full; \scr{} agrees with full
recomputation on \agreeTScr\% of questions and prefills \tokSaveT\% fewer tokens. Its selection contains the needle
segment on \locRecallT\% of questions (\locRecallUnseenT\% when the sparse pass missed it).
On LongVideoBench, \scr{} agrees with full recomputation on \agreeScrLVBQ\% (Qwen3-VL-8B) and \agreeScrLVBT\%
(Thinking) of questions and prefills \tokSaveLVBQ\% and \tokSaveLVBT\% fewer tokens. The latency gain is smaller than
for Qwen2.5-VL-7B (\speedLVBQ$\times$ and \speedLVBT$\times$, \cref{tab:models}) because Qwen3-VL's full rebuild is
already faster (\latFullLVBQ{}\,s vs.\ \latFullLVB{}\,s), so the ViT time that \scr{} skips is a smaller share.

\begin{table*}[t]
  \centering
  \caption{\textbf{Three models.} Accuracy (\%) on V-NIAH (720 questions), Video-MME long (180) and LongVideoBench
  ($\geq$20 min, 276), and \scr{} answer agreement with, prefill saving (\%) over and rebuild speedup over full
  recomputation (SDPA).}
  \label{tab:models}
  \small
  \resizebox{\textwidth}{!}{\IfFileExists{tables/tab_models.tex}{\begin{tabular}{lccccccccc}
\toprule
 & \multicolumn{3}{c}{Qwen2.5-VL-7B} & \multicolumn{3}{c}{Qwen3-VL-8B} & \multicolumn{3}{c}{Qwen3-VL-8B-Thinking} \\
\cmidrule(lr){2-4}\cmidrule(lr){5-7}\cmidrule(lr){8-10}
Condition & V-NIAH & Video-MME & LVB & V-NIAH & Video-MME & LVB & V-NIAH & Video-MME & LVB \\
\midrule
uniform (no edit) & 52.2 & 55.6 & 56.2 & 55.6 & 60.0 & 55.1 & 52.8 & 58.9 & 59.1 \\
edit + full recompute & 49.6 & 53.9 & 55.4 & 53.3 & 61.7 & 56.9 & 52.4 & 55.6 & 56.9 \\
edit + prefix reuse & 49.6 & 54.4 & 56.2 & 53.6 & 62.8 & 57.6 & 52.6 & 56.7 & 56.9 \\
edit + SCR-VL (ours) & 50.7 & 54.4 & 55.1 & 54.4 & 62.8 & 56.2 & 52.4 & 57.2 & 56.9 \\
\midrule
oracle edit + full & 96.9 & -- & -- & 96.8 & -- & -- & 96.8 & -- & -- \\
oracle edit + SCR-VL & 96.9 & -- & -- & 96.8 & -- & -- & 96.8 & -- & -- \\
\midrule
SCR-VL answer agreement & 97.4 & 96.1 & 97.1 & 95.8 & 96.7 & 98.9 & 97.2 & 97.2 & 94.6 \\
SCR-VL prefill saving & 30 & 32 & 32 & 34 & 30 & 29 & 31 & 43 & 33 \\
SCR-VL rebuild speedup ($\times$) & 1.7 & 1.8 & 1.8 & 1.3 & 1.2 & 1.2 & 1.2 & 1.5 & 1.2 \\
\bottomrule
\end{tabular}
}{(not available)}}
\end{table*}

\section{Limitations}
\label{sec:limits}
These are \emph{preliminary} results. (1) \scr{} is approximate beyond layer 0; we measure its effect on the answer
but not on long generations. (2) Three models from one family (Qwen2.5-VL-7B, Qwen3-VL-8B Instruct and Thinking, the latter without reasoning), one edit rule (densify $\le$3 selected, delete even
unselected) and one prompt. (3) The main benchmark is synthetic: the needle is an inserted title card, easier to
recognise than natural events. (4) Latencies come from a HuggingFace implementation on a shared machine, not a serving
system; with SDPA the LLM-side time of \scr{} was \emph{not} lower than full recomputation (with FlashAttention-2 it
was \llmSaveFA\% lower), and much of the end-to-end gain comes from the ViT cache. (5) The real-video samples are
small: \nMME{} Video-MME questions on \nMMEvid{} videos and \nLVB{} LongVideoBench questions on \nLVBvid{} videos; MLVU
was not run. The answer-free study of larger $K$ (\cref{tab:infra}) uses \nInfra{} questions.
(6) The ablation subset contains only 20-minute haystacks. (7) The model's own localization does not improve accuracy
here, and at a 64K budget it collapses to a single segment; whether training or better prompting closes the gap to the oracle edit is open. (8) Within our 32K budget the appended context always fits, and appending beats in-place
editing; the regime where segments must be deleted to stay within budget, which motivates in-place editing and \scr{},
is not tested.

\section{Conclusion}
Moving a video segment shifts all three M-RoPE components by the same amount, so the cached keys of surviving segments
can be reused after a mid-context edit with one rotation, and their ViT outputs can be reused exactly. On our tests this
keeps answers close to full recomputation while cutting prefill tokens by \tokSaveFull\% and rebuild latency by
\latSaveFull\%. The edited context itself can nearly solve hour-long needle search when the right segment is chosen;
getting the 7--8B VLMs we tested to choose it is the open problem. Within a 32K budget, appending the densified
segments was both more accurate and cheaper than editing in place, so the case for \scr{} rests on budgets where
deletion is required.

\bibliography{refs}

@article{yao2024cacheblend,
  title         = {{CacheBlend: Fast Large Language Model Serving for RAG with Cached Knowledge Fusion}},
  author        = {Yao, Jiayi and Li, Hanchen and Liu, Yuhan and Ray, Siddhant and Cheng, Yihua and Zhang, Qizheng and Du, Kuntai and Lu, Shan and Jiang, Junchen},
  year          = {2024},
  eprint        = {2405.16444},
  journal       = {arXiv preprint arXiv:2405.16444},
  archivePrefix = {arXiv}
}

@inproceedings{gim2024promptcache,
  title     = {{Prompt Cache: Modular Attention Reuse for Low-Latency Inference}},
  author    = {Gim, In and Chen, Guojun and Lee, Seung-seob and Sarda, Nikhil and Khandelwal, Anurag and Zhong, Lin},
  booktitle = {{MLSys}},
  year      = {2024},
  eprint    = {2311.04934},
  archivePrefix = {arXiv}
}

@article{hu2024epic,
  title         = {{EPIC: Efficient Position-Independent Caching for Serving Large Language Models}},
  author        = {Hu, Junhao and Huang, Wenrui and Wang, Weidong and Wang, Haoyi and Hu, Tiancheng and Zhang, Qin and Feng, Hao and Chen, Xusheng and Shan, Yizhou and Xie, Tao},
  year          = {2024},
  eprint        = {2410.15332},
  journal       = {arXiv preprint arXiv:2410.15332},
  archivePrefix = {arXiv}
}

@article{zheng2023sglang,
  title         = {{SGLang: Efficient Execution of Structured Language Model Programs}},
  author        = {Zheng, Lianmin and Yin, Liangsheng and Xie, Zhiqiang and Sun, Chuyue and Huang, Jeff and Yu, Cody Hao and Cao, Shiyi and Kozyrakis, Christos and Stoica, Ion and Gonzalez, Joseph E. and Barrett, Clark and Sheng, Ying},
  year          = {2023},
  eprint        = {2312.07104},
  journal       = {arXiv preprint arXiv:2312.07104},
  archivePrefix = {arXiv}
}

@inproceedings{xiao2024streamingllm,
  title     = {{Efficient Streaming Language Models with Attention Sinks}},
  author    = {Xiao, Guangxuan and Tian, Yuandong and Chen, Beidi and Han, Song and Lewis, Mike},
  booktitle = {{ICLR}},
  year      = {2024},
  eprint    = {2309.17453},
  archivePrefix = {arXiv}
}

@inproceedings{di2025rekv,
  title     = {{Streaming Video Question-Answering with In-context Video KV-Cache Retrieval}},
  author    = {Di, Shangzhe and Yu, Zhelun and Zhang, Guanghao and Li, Haoyuan and Zhong, Tao and Cheng, Hao and Li, Bolin and He, Wanggui and Shu, Fangxun and Jiang, Hao},
  booktitle = {{ICLR}},
  year      = {2025},
  eprint    = {2503.00540},
  archivePrefix = {arXiv}
}

@inproceedings{xu2026streamingvlm,
  title     = {{StreamingVLM: Real-Time Understanding for Infinite Video Streams}},
  author    = {Xu, Ruyi and Xiao, Guangxuan and Chen, Yukang and He, Liuning and Lu, Yao and Han, Song},
  booktitle = {{ICLR}},
  year      = {2026},
  eprint    = {2510.09608},
  archivePrefix = {arXiv}
}

@article{shu2024videoxl,
  title         = {{Video-XL: Extra-Long Vision Language Model for Hour-Scale Video Understanding}},
  author        = {Shu, Yan and Liu, Zheng and Zhang, Peitian and Qin, Minghao and Zhou, Junjie and Liang, Zhengyang and Huang, Tiejun and Zhao, Bo},
  year          = {2024},
  eprint        = {2409.14485},
  journal       = {arXiv preprint arXiv:2409.14485},
  archivePrefix = {arXiv}
}

@article{qin2025videoxl2,
  title         = {{Video-XL-2: Towards Very Long-Video Understanding Through Task-Aware KV Sparsification}},
  author        = {Qin, Minghao and Liu, Xiangrui and Liang, Zhengyang and Shu, Yan and Yuan, Huaying and Zhou, Juenjie and Xiao, Shitao and Zhao, Bo and Liu, Zheng},
  year          = {2025},
  eprint        = {2506.19225},
  journal       = {arXiv preprint arXiv:2506.19225},
  archivePrefix = {arXiv}
}

@article{zhang2024longva,
  title         = {{Long Context Transfer from Language to Vision}},
  author        = {Zhang, Peiyuan and Zhang, Kaichen and Li, Bo and Zeng, Guangtao and Yang, Jingkang and Zhang, Yuanhan and Wang, Ziyue and Tan, Haoran and Li, Chunyuan and Liu, Ziwei},
  year          = {2024},
  eprint        = {2406.16852},
  journal       = {arXiv preprint arXiv:2406.16852},
  archivePrefix = {arXiv}
}

@article{bai2025qwen25vl,
  title         = {{Qwen2.5-VL Technical Report}},
  author        = {Bai, Shuai and Chen, Keqin and Liu, Xuejing and Wang, Jialin and Ge, Wenbin and Song, Sibo and Dang, Kai and Wang, Peng and Wang, Shijie and Tang, Jun and Zhong, Humen and Zhu, Yuanzhi and Yang, Mingkun and Li, Zhaohai and Wan, Jianqiang and Wang, Pengfei and Ding, Wei and Fu, Zheren and Xu, Yiheng and Ye, Jiabo and Zhang, Xi and Xie, Tianbao and Cheng, Zesen and Zhang, Hang and Yang, Zhibo and Xu, Haiyang and Lin, Junyang},
  year          = {2025},
  eprint        = {2502.13923},
  journal       = {arXiv preprint arXiv:2502.13923},
  archivePrefix = {arXiv}
}

@article{wang2024qwen2vl,
  title         = {{Qwen2-VL: Enhancing Vision-Language Model's Perception of the World at Any Resolution}},
  author        = {Wang, Peng and Bai, Shuai and Tan, Sinan and Wang, Shijie and Fan, Zhihao and Bai, Jinze and Chen, Keqin and Liu, Xuejing and Wang, Jialin and Ge, Wenbin and Fan, Yang and Dang, Kai and Du, Mengfei and Ren, Xuancheng and Men, Rui and Liu, Dayiheng and Zhou, Chang and Zhou, Jingren and Lin, Junyang},
  year          = {2024},
  eprint        = {2409.12191},
  journal       = {arXiv preprint arXiv:2409.12191},
  archivePrefix = {arXiv}
}

@article{bai2025qwen3vl,
  title         = {{Qwen3-VL Technical Report}},
  author        = {Bai, Shuai and Cai, Yuxuan and Chen, Ruizhe and Chen, Keqin and Chen, Xionghui and Cheng, Zesen and Deng, Lianghao and Ding, Wei and Gao, Chang and Ge, Chunjiang and Ge, Wenbin and Guo, Zhifang and Huang, Qidong and Huang, Jie and Huang, Fei and Hui, Binyuan and Jiang, Shutong and Li, Zhaohai and Li, Mingsheng and Li, Mei and Li, Kaixin and Lin, Zicheng and Lin, Junyang and Liu, Xuejing and Liu, Jiawei and Liu, Chenglong and Liu, Yang and Liu, Dayiheng and Liu, Shixuan and Lu, Dunjie and Luo, Ruilin and Lv, Chenxu and Men, Rui and Meng, Lingchen and Ren, Xuancheng and Ren, Xingzhang and Song, Sibo and Sun, Yuchong and Tang, Jun and Tu, Jianhong and Wan, Jianqiang and Wang, Peng and Wang, Pengfei and Wang, Qiuyue and Wang, Yuxuan and Xie, Tianbao and Xu, Yiheng and Xu, Haiyang and Xu, Jin and Yang, Zhibo and Yang, Mingkun and Yang, Jianxin and Yang, An and Yu, Bowen and Zhang, Fei and Zhang, Hang and Zhang, Xi and Zheng, Bo and Zhong, Humen and Zhou, Jingren and Zhou, Fan and Zhou, Jing and Zhu, Yuanzhi and Zhu, Ke},
  year          = {2025},
  eprint        = {2511.21631},
  journal       = {arXiv preprint arXiv:2511.21631},
  archivePrefix = {arXiv}
}

@article{fu2024videomme,
  title         = {{Video-MME: The First-Ever Comprehensive Evaluation Benchmark of Multi-modal LLMs in Video Analysis}},
  author        = {Fu, Chaoyou and Dai, Yuhan and Luo, Yongdong and Li, Lei and Ren, Shuhuai and Zhang, Renrui and Wang, Zihan and Zhou, Chenyu and Shen, Yunhang and Zhang, Mengdan and Chen, Peixian and Li, Yanwei and Lin, Shaohui and Zhao, Sirui and Li, Ke and Xu, Tong and Zheng, Xiawu and Chen, Enhong and Shan, Caifeng and He, Ran and Sun, Xing},
  year          = {2024},
  eprint        = {2405.21075},
  journal       = {arXiv preprint arXiv:2405.21075},
  archivePrefix = {arXiv}
}

@article{zhou2024mlvu,
  title         = {{MLVU: Benchmarking Multi-task Long Video Understanding}},
  author        = {Zhou, Junjie and Shu, Yan and Zhao, Bo and Wu, Boya and Liang, Zhengyang and Xiao, Shitao and Qin, Minghao and Yang, Xi and Xiong, Yongping and Zhang, Bo and Huang, Tiejun and Liu, Zheng},
  year          = {2024},
  eprint        = {2406.04264},
  journal       = {arXiv preprint arXiv:2406.04264},
  archivePrefix = {arXiv}
}

@article{wu2024longvideobench,
  title         = {{LongVideoBench: A Benchmark for Long-context Interleaved Video-Language Understanding}},
  author        = {Wu, Haoning and Li, Dongxu and Chen, Bei and Li, Junnan},
  year          = {2024},
  eprint        = {2407.15754},
  journal       = {arXiv preprint arXiv:2407.15754},
  archivePrefix = {arXiv}
}

@article{wang2024videoagent,
  title         = {{VideoAgent: Long-form Video Understanding with Large Language Model as Agent}},
  author        = {Wang, Xiaohan and Zhang, Yuhui and Zohar, Orr and Yeung-Levy, Serena},
  year          = {2024},
  eprint        = {2403.10517},
  journal       = {arXiv preprint arXiv:2403.10517},
  archivePrefix = {arXiv}
}

@inproceedings{wang2025videotree,
  title     = {{VideoTree: Adaptive Tree-based Video Representation for LLM Reasoning on Long Videos}},
  author    = {Wang, Ziyang and Yu, Shoubin and Stengel-Eskin, Elias and Yoon, Jaehong and Cheng, Feng and Bertasius, Gedas and Bansal, Mohit},
  booktitle = {{CVPR}},
  year      = {2025},
  eprint    = {2405.19209},
  archivePrefix = {arXiv}
}

@inproceedings{zhang2025deepvideodiscovery,
  title     = {{Deep Video Discovery: Agentic Search with Tool Use for Long-form Video Understanding}},
  author    = {Zhang, Xiaoyi and Jia, Zhaoyang and Guo, Zongyu and Li, Jiahao and Li, Bin and Li, Houqiang and Lu, Yan},
  booktitle = {{NeurIPS}},
  year      = {2025},
  eprint    = {2505.18079},
  archivePrefix = {arXiv}
}

@article{su2021roformer,
  title         = {{RoFormer: Enhanced Transformer with Rotary Position Embedding}},
  author        = {Su, Jianlin and Lu, Yu and Pan, Shengfeng and Murtadha, Ahmed and Wen, Bo and Liu, Yunfeng},
  year          = {2021},
  eprint        = {2104.09864},
  journal       = {arXiv preprint arXiv:2104.09864},
  archivePrefix = {arXiv}
}

@inproceedings{kwon2023pagedattention,
  title     = {{Efficient Memory Management for Large Language Model Serving with PagedAttention}},
  author    = {Kwon, Woosuk and Li, Zhuohan and Zhuang, Siyuan and Sheng, Ying and Zheng, Lianmin and Yu, Cody Hao and Gonzalez, Joseph E. and Zhang, Hao and Stoica, Ion},
  booktitle = {{SOSP}},
  year      = {2023},
  eprint    = {2309.06180},
  archivePrefix = {arXiv}
}

@article{shao2026clm,
  title         = {{Context Language Models}},
  author        = {Shao, Rulin and Shen, Shannon Zejiang and Yin, Junjie Oscar and Li, Yuetai and Wang, Minheng and Ivison, Hamish and Poovendran, Radha and Lambert, Nathan and Xiao, Teng and Lewis, Mike and Yih, Wen-tau and Zettlemoyer, Luke and Koh, Pang Wei},
  year          = {2026},
  eprint        = {2609.37725},
  journal       = {arXiv preprint arXiv:2609.37725},
  archivePrefix = {arXiv},
  url           = {https://github.com/facebookresearch/context-language-models}
}
\bibliographystyle{icml2026}

\newpage
\appendix
\onecolumn

\section{Implementation details}
\label{app:impl}
\textbf{Blocks.} The context is a list of blocks (system text, segments, question, prompts), each with a content hash;
a segment block is its label tokens, \texttt{<|vision\_start|>}, the video pad tokens, \texttt{<|vision\_end|>} and a
newline. Each segment's frames are processed by the model's image processor and encoded by the ViT on their own, so
the ViT output of a segment does not depend on its neighbours. Position ids always come from the model's
\texttt{get\_rope\_index} on the full token sequence, with per-segment \texttt{second\_per\_grid\_ts}$=2/\text{fps}$.

\textbf{Cache.} transformers 4.51 \texttt{DynamicCache} stores per-layer keys and values of shape
$[1, 4, L, 128]$ with keys already rotated. Relocation slices the old cache, rotates keys by \cref{eq:shift} in fp32,
casts to bf16 and appends with \texttt{cache.update}. Prefill on top of a cache passes explicit \texttt{position\_ids}
of shape $[3,1,L]$ and \texttt{cache\_position}. Answer and localization prompts are appended to a built cache and
cropped off afterwards.

\textbf{Frames.} Films are decoded once into a 2\,fps frame bank at $448{\times}308$; Video-MME videos into a 1\,fps
bank at $448{\times}252$. A segment samples frames at the centres of $n_i$ equal sub-intervals. Needle frames replace a
sampled frame when its timestamp falls inside the needle interval.

\textbf{Hyperparameters.} $N{=}32$ segments; 8 initial and 32 densified frames per segment; at most 3 selected
segments; $K{=}6$, $r{=}0$ by default; greedy decoding with at most 24 new tokens for localization; seed 0.

\section{Correctness tests}
\label{app:tests}
\texttt{tests/test\_correctness.py} runs on a 12-segment, 4-frame context with an oracle edit at 16 frames; all pass
(\texttt{5 passed}).
\begin{enumerate}
\item \textbf{Position consistency.} For \testSegs{} segment blocks in the old and new contexts, the positions from
\texttt{get\_rope\_index} on the full sequence equal the segment's local positions plus one constant; every survivor
has a single $\Delta$ over $(t,h,w)$.
\item \textbf{Layer-0 exactness.} Relocated layer-0 keys vs.\ a full prefill of the new sequence: max relative error
\testRelErr{}, max absolute error \testAbsErr{} (max key magnitude \testMaxKey{}); without rotation the smallest
relative error is \testNoShift{}. Layer-0 values are bitwise equal.
\item \textbf{No-edit equivalence.} \scr{} vs.\ full on an unchanged context: KL \testNoEditKL{}, same top-1 token.
\item \textbf{Prefix exactness.} prefix vs.\ full: KL \testPrefixKL{}, same top-1 token.
\item \textbf{Tail deletion.} Deleting only the last segment: \scr{} relocates nothing and its logits equal prefix
reuse bitwise.
\end{enumerate}

\section{Additional results}
\label{app:more}

\begin{table}[h]
  \centering
  \caption{Ablation of $K$ and $r$ on the \nAbl{}-question subset (all 20-minute haystacks). KL and agreement are
  relative to full recomputation on the same edited context.}
  \label{tab:ablation}
  \small
  \begin{tabular}{lrrrrr}
\toprule
Setting & prefill tok. & latency (s) & KL & answer agree & acc. \\
\midrule
full recompute & 17{,}556 & -- & 0 & 100.0\% & 70.4 \\
$K{=}1{,} r{=}0$ & 16{,}053 & 4.96 & 0.0009 & 98.8\% & 70.8 \\
$K{=}3{,} r{=}0$ & 14{,}587 & 4.70 & 0.0061 & 97.9\% & 70.8 \\
$K{=}6{,} r{=}0$ & 12{,}388 & 4.37 & 0.0082 & 96.7\% & 71.2 \\
$K{=}12{,} r{=}0$ & 7{,}992 & 3.86 & 0.0142 & 96.7\% & 71.2 \\
$K{=}6{,} r{=}16$ & 12{,}484 & 4.61 & 0.0079 & 97.1\% & 71.2 \\
$K{=}6{,} r{=}64$ & 12{,}772 & 4.63 & 0.0082 & 97.1\% & 71.2 \\
\bottomrule
\end{tabular}

\end{table}

\begin{table}[h]
  \centering
  \caption{Initial budget 32K vs.\ 64K on the same \nBudget{} V-NIAH questions (densified segments: 32 vs.\ 64
  frames). Rebuild tokens are prefilled tokens after the edit.}
  \label{tab:budget}
  \small
  \IfFileExists{tables/tab_budget.tex}{\begin{tabular}{lrrrr}
\toprule
Initial frames / segment & \multicolumn{2}{c}{8 (32K budget)} & \multicolumn{2}{c}{16 (64K budget)} \\
\cmidrule(lr){2-3}\cmidrule(lr){4-5}
 & acc. (\%) & rebuild tok. & acc. (\%) & rebuild tok. \\
\midrule
initial context & -- & 23{,}549 & -- & 46{,}077 \\
uniform (no edit) & 52.2\,{\scriptsize[49,56]} & -- & 63.1\,{\scriptsize[60,67]} & -- \\
edit + full recompute & 49.6\,{\scriptsize[46,53]} & 17{,}361 & 53.9\,{\scriptsize[50,57]} & 29{,}511 \\
edit + prefix reuse & 49.6\,{\scriptsize[46,53]} & 16{,}571 & 54.2\,{\scriptsize[51,58]} & 28{,}031 \\
edit + SCR-VL (ours) & 50.7\,{\scriptsize[47,54]} & 12{,}191 & 56.1\,{\scriptsize[52,60]} & 19{,}409 \\
oracle edit + full & 96.9\,{\scriptsize[96,98]} & 14{,}455 & 100.0\,{\scriptsize[100,100]} & 28{,}317 \\
oracle edit + SCR-VL & 96.9\,{\scriptsize[96,98]} & 9{,}281 & 100.0\,{\scriptsize[100,100]} & 18{,}215 \\
needle sampled initially & 37.1 & & 67.8 & \\
selection hits needle & 29.2 & & 7.5 & \\
\bottomrule
\end{tabular}
}{(not available)}
\end{table}

\begin{table}[h]
  \centering
  \caption{Cost per question on Video-MME long (\nMME{} questions); columns as in \cref{tab:cost}.}
  \label{tab:cost_mme}
  \small
  \IfFileExists{tables/tab_cost_mme.tex}{\begin{tabular}{lrrrrrrr}
\toprule
Rebuild path & prefill & moved & ViT & TFLOPs & latency & LLM-only & mem \\
 & tokens & tokens & frames & & (s) & (s) & (GB) \\
\midrule
initial context & 19{,}461 & -- & 256 & 472 & 7.68 & 1.70 & 20.0 \\
\midrule
full recompute & 14{,}800 & 0 & 197 & 346 & 6.30 & 1.21 & 19.4 \\
prefix reuse & 13{,}502 & 0 & 180 & 317 & 6.59 & 1.90 & 20.4 \\
SCR-VL & 10{,}013 & 3{,}489 & 71 & 211 & 3.46 & 1.31 & 19.1 \\
\bottomrule
\end{tabular}
}{(not available)}
\end{table}

\begin{table}[h]
  \centering
  \caption{Cost per question on LongVideoBench ($\geq$20 min, \nLVB{} questions, Qwen2.5-VL-7B); columns as in
  \cref{tab:cost}.}
  \label{tab:cost_lvb}
  \small
  \IfFileExists{tables/tab_cost_lvb.tex}{\begin{tabular}{lrrrrrrr}
\toprule
Rebuild path & prefill & moved & ViT & TFLOPs & latency & LLM-only & mem \\
 & tokens & tokens & frames & & (s) & (s) & (GB) \\
\midrule
initial context & 19{,}515 & -- & 256 & 473 & 7.60 & 1.75 & 20.0 \\
\midrule
full recompute & 14{,}851 & 0 & 197 & 348 & 6.23 & 1.23 & 19.4 \\
prefix reuse & 13{,}640 & 0 & 181 & 320 & 6.57 & 1.93 & 20.4 \\
SCR-VL & 10{,}129 & 3{,}512 & 72 & 214 & 3.49 & 1.33 & 19.1 \\
\bottomrule
\end{tabular}
}{(not available)}
\end{table}

\section{Answer-free evaluation of rebuild paths}
\label{app:infra}
\textbf{Protocol.} The edits of the frozen edit rule are recorded once per question (old and new segment lists) and
replayed through every rebuild path, so all paths see byte-identical edits. Full recomputation is the reference. For
each case we let the exact cache greedily generate 32 tokens after a neutral prompt that opens an assistant turn, and
teacher-force that continuation through every candidate cache. \emph{Probe KL} is the mean per-position KL(full$\,\|\,$path)
and \emph{argmax agree} the fraction of positions with the same argmax token. Prefix reuse is exact in exact arithmetic,
so its error measures bf16 chunking noise. Latency is the fastest of 2 repeats. All rows use Qwen2.5-VL-7B on the fixed
\nInfra{}-question V-NIAH sample (10 per film and haystack length).

\textbf{Real videos.} On fixed samples of Video-MME (\nInfraMME{} questions) and LongVideoBench (\nInfraLVB{}),
relocating all survivors saves \infraTokAllMME\% and \infraTokAllLVB\% of prefill tokens at \infraSpeedAllMME$\times$
and \infraSpeedAllLVB$\times$ (SDPA), with probe KL \infraKLAllMME{} and \infraKLAllLVB{} (vs.\ \infraKLSixMME{} and
\infraKLSixLVB{} at $K{=}6$). The fidelity cost of large $K$ is therefore somewhat higher on real videos than on V-NIAH.

\begin{table}[h]
  \centering
  \caption{\textbf{Rebuild paths on the answer-free harness} (\nInfra{} V-NIAH questions, Qwen2.5-VL-7B). Savings and
  speedups are relative to full recomputation on the same edit; FlashAttention-2 was run for the default and the
  all-survivor setting only.}
  \label{tab:infra}
  \small
  \IfFileExists{tables/tab_infra.tex}{\begin{tabular}{lrrrrrrrr}
\toprule
 & prefill & TFLOPs & \multicolumn{2}{c}{SDPA} & \multicolumn{2}{c}{FlashAttention-2} & probe & probe \\
\cmidrule(lr){4-5}\cmidrule(lr){6-7}
Rebuild path & saving & saving & latency (s) & speedup & latency (s) & speedup & KL & argmax agree \\
\midrule
full recompute & 0\% & 0\% & 7.22 & 1.00 & 3.49 & 1.00 & 0 & 100.0\% \\
prefix reuse & 4\% & 4\% & 8.14 & 0.89 & 3.37 & 1.04 & 0.0012 & 98.8\% \\
SCR-VL, $K{=}3$ & 17\% & 26\% & 4.62 & 1.56 & -- & -- & 0.0031 & 98.7\% \\
SCR-VL, $K{=}6$ (default) & 30\% & 36\% & 4.29 & 1.68 & 1.90 & 1.84 & 0.0081 & 97.8\% \\
SCR-VL, $K{=}12$ & 55\% & 57\% & 3.78 & 1.91 & -- & -- & 0.0210 & 95.1\% \\
SCR-VL, all survivors & 63\% & 63\% & 3.65 & 1.98 & 1.52 & 2.29 & 0.0242 & 94.9\% \\
\bottomrule
\end{tabular}
}{(not available)}
\end{table}

\section{Compute and data}
\label{app:data}
\textbf{Compute.} One 8$\times$A800-80GB machine shared with other jobs. Summed per-question wall-clock is
\gpuHoursVniah{} GPU-hours for V-NIAH, \gpuHoursBudget{} for the 64K-budget run, \gpuHoursMME{} for Video-MME,
\gpuHoursFA{} for the FlashAttention-2 re-run, \gpuHoursAppend{} for edit-append, and \gpuHoursQ{}/\gpuHoursQMME{}
(Qwen3-VL-8B) and \gpuHoursT{}/\gpuHoursTMME{} (Qwen3-VL-8B-Thinking) for V-NIAH/Video-MME, and
\gpuHoursLVB{}/\gpuHoursLVBQ{}/\gpuHoursLVBT{} for LongVideoBench with the three models; model download, frame decoding and tests are
not included.

\begin{table}[h]
  \centering
  \caption{Data sources and licenses. No video is redistributed; the V-NIAH generator and frame-bank scripts recreate
  the inputs from the original files.}
  \label{tab:data}
  \small
  \resizebox{\textwidth}{!}{%
  \begin{tabular}{llll}
    \toprule
    Source & Identifier & License / terms & Use \\
    \midrule
    \emph{Anna Christie} (1930) & archive.org \texttt{anna-christie-1930-greta-garbo-english} & Public Domain Mark 1.0 & V-NIAH haystack \\
    \emph{Disraeli} (1929) & archive.org \texttt{disraeli\_1929\_202506} & Public Domain Mark 1.0 & V-NIAH haystack \\
    \emph{Bulldog Drummond} (1929) & archive.org \texttt{bulldog-drummond\_1929\_202506} & Public Domain Mark 1.0 & V-NIAH haystack \\
    Video-MME & HF \texttt{lmms-lab/Video-MME} (first 60 long videos) & academic research use & real-video QA \\
    LongVideoBench & HF \texttt{longvideobench/LongVideoBench} (val, videos $\geq$20 min) & CC BY-NC-SA 4.0 & real-video QA \\
    Qwen2.5-VL-7B-Instruct & HF \texttt{Qwen/Qwen2.5-VL-7B-Instruct} & Apache-2.0 & model \\
    Qwen3-VL-8B Instruct / Thinking & HF \texttt{Qwen/Qwen3-VL-8B-\{Instruct,Thinking\}} & Apache-2.0 & model \\
    \bottomrule
  \end{tabular}}
\end{table}

\end{document}